# NourishNet: Proactive Severity State Forecasting of Food Commodity Prices for Global Warning Systems

Sydney Balboni, Grace Ivey, Brett Storoe, John Cisler, Tyge Plater, Caitlyn Grant, Ella Bruce, Benjamin Paulson*

Department of Electrical Engineering and Computer Science

Milwaukee School of Engineering

1025 N Broadway St, Milwaukee, WI 53202

balbonis@msoe.edu, iveyg@msoe.edu, storoeb@msoe.edu, cislerj@msoe.edu, platert@msoe.edu, grantc@msoe.edu, brucee@msoe.edu, paulsonb@msoe.edu

## Abstract

Price volatility in global food commodities is a critical signal indicating potential disruptions in the food market. Understanding forthcoming changes in these prices is essential for bolstering food security, particularly for nations at risk. The Food and Agriculture Organization of the United Nations (FAO) previously developed sophisticated statistical frameworks for the proactive prediction of food commodity prices, aiding in the creation of global early warning systems. These frameworks utilize food security indicators to produce accurate forecasts, thereby facilitating preparations against potential food shortages. Our research builds on these foundations by integrating robust price security indicators with cutting-edge deep learning (DL) methodologies to reveal complex interdependencies. DL techniques examine intricate dynamics among diverse factors affecting food prices. Through sophisticated time-series forecasting models coupled with a classification model, our approach enhances existing models to better support communities worldwide in advancing their food security initiatives.

# 1 Introduction

In the world of volatile food commodity markets, accurate price forecasting remains a crucial tool for consumers to protect themselves from rapid price increases. Lower income countries spend a significant proportion of their income on food commodities. For instance, Nigerian communities spend almost 60% of total household funds on food expenditures according to USDA statistics [1] and at least 690 million people faced hunger in 2022 [2]. The ability to anticipate price fluctuations holds profound implications for economic stability, food security, and investment decisions.

Traditional approaches to market forecasting rely heavily on historical commodity data and market observations to report warnings associated with abnormalities in price trends. The Global Information and Early Warning System on Food and Agriculture (GIEWS) [3] is the solution curated by our affiliate, the Food and Agriculture Organization of the United Nations (FAO). GIEWS describes anomalous food supply and agricultural situations in countries and subregions. However, the increasing complexity and interconnectedness of global markets call for more sophisticated analytical tools to complement existing methods and capture the multifaceted drivers of price changes.

Our solution introduces a novel DL approach to food commodity price forecasting, leveraging the Proteus Index [4], an array of historical prices of commodities, food price indices, and futures, in conjunction with sentiment analysis derived from financial news. At the heart of our methodology is the application of a transformer network [5], which demonstrates exceptional efficacy in capturing dependencies within time-series to create accurate predictions [6]. Additionally, incorporating sentiment analysis enhances the model's ability to assess market status and refine its predictions – a transformer from Hugging Face [7] was used and fine-tuned to accomplish the sentiment analysis.

Our research objective is to present a more accurate model for predicting food commodity prices that demonstrates the potential of advanced DL techniques in enhancing our understanding of market dynamics. Through the application of this multifaceted approach, we seek to offer a valuable tool to augment current informed decision-making processes in the face of market volatility. The official implementation does not encompass end-to-end decision-making; instead, it serves as advisory input for the final decisions.

# 2 Data

## 2.1 Data Collection

A collection of monthly market data obtained from an assortment of repositories, each yielded through different methodologies, is used in the final intake process for inferencing price and warning predictions. This collection presents its own challenges – as outlined in section 5.2, 'Data Overlap Scarcity' – though the data is ultimately aggregated into a single, comprehensive dataset.

Of the final dataset employed for prediction, the following features are included from the FAO: Proteus index [8], an encoded measurement of a country's food security [4]; food commodity price by country [9] to capture regional variations in price; FAO commodity price indices [10] to capture global variations in price; harvest data [11] as an indicator for shortages or surpluses [12]; and annual FAO Agriculture Outlook reports [13] to capture limited market demand. Alongside this subsection of the final data features, the following features are included from other sources: food commodity futures [14] to capture spot price outlook [15]; and international financial statistics (IFS) [16] for global financial comparison. Consequently, this meticulously curated dataset offers a multifaceted perspective of the food commodity market through seven distinct features, serving as the backbone for both the price-prediction and warning-prediction models, with the latter necessitating manual extraction of FAO warnings [17] as labels, a process detailed in section 5.1, 'FAO Warning Labels'.

## 2.2 Data Cleaning

While a variety of datapoints are employed in different portions of the overall pipeline – as outlined in section 3, 'Methods' – the data compiled for processing goes through a standard cleaning process to ensure datapoints were consistent for both the price-prediction and warning-prediction DL models. This cleaning involves normalization of all numerical values between zero and one (inclusive), and a detrending process of any time-series datapoints. Because prior work in the space of increasing time-series prediction model capabilities has supported the idea that normalization aids in the predictive capabilities of the model [18], normalization is included in the data cleaning process. Similarly, previous work supports the use of detrending in time-series prediction applications as a method of ensuring data consistency throughout lengthy time ranges and market fluctuations [19]. Beyond these noteworthy approaches, typical cleaning steps for DL applications – such as standard scaling [20] and tabularly-formatted data for transformer architectures [21] – were also employed. Collectively, these data cleaning procedures are fundamental in enhancing the reliability of our price-prediction and warning-prediction DL models.

# 3 Methods

## 3.1 Model Development

The overall model, NourishNet, is developed in the context of increasing food security understanding for underdeveloped communities, as well as providing accurate yet early insights into which food commodity markets will soon experience a price spike and therefore result in food scarcity in relevant markets. The specific implementation can be found on our GitHub [22], and an overview of the full NourishNet pipeline is further described in the following sections.

A high-level overview of the NourishNet pipeline is as follows: data from a variety of data repositories is cleaned via a standardized process, outlined in section 3.2; cleaned data is passed to a time-series prediction model for food commodity price over a variable number of months, as outlined in section 3.1.1; results from previous steps are used to predict warnings following output guidelines of GIEWS, as outlined in 3.1.2; and an LLM is used to provide data insights to users lacking an understanding market dynamics, as outlined in section 3.3.

### 3.1.1 Predicting Food Commodity Price

Price prediction remains a critical area of interest as its successful implementation provides valuable insights into market dynamics that facilitate informed decisions regarding monetary allocation across a variety of markets. This model provides state-of-the-art prediction capabilities compared to publicly available prior work [23] in the domain of predicting food commodity prices across $n$ months. These predictions are integral to predicting when the FAO will send out a warning based on current market features and the predicted market price-state, as outlined in section 3.1.2.

The architecture of this model follows the traditional encoder-only transformer [5]. However, some architectural changes were made to provide prediction improvements for the data domain. These changes involve the inclusion of multiple transformer blocks to enhance the model's capacity for complex pattern recognition [24], using the following new layers: batch normalization after the feed-forward network (FFN) is used to stabilize the learning process [25]; dropout in the FFN to reduce the likelihood of overfitting [26]; and global average pooling to guarantee uniform output dimensions irrespective of the diversity of data sources. The specific data format is outlined in section 4.1, 'Formatting Input Data' and the full architecture of the encoder-only transformer is depicted in Figure 1.

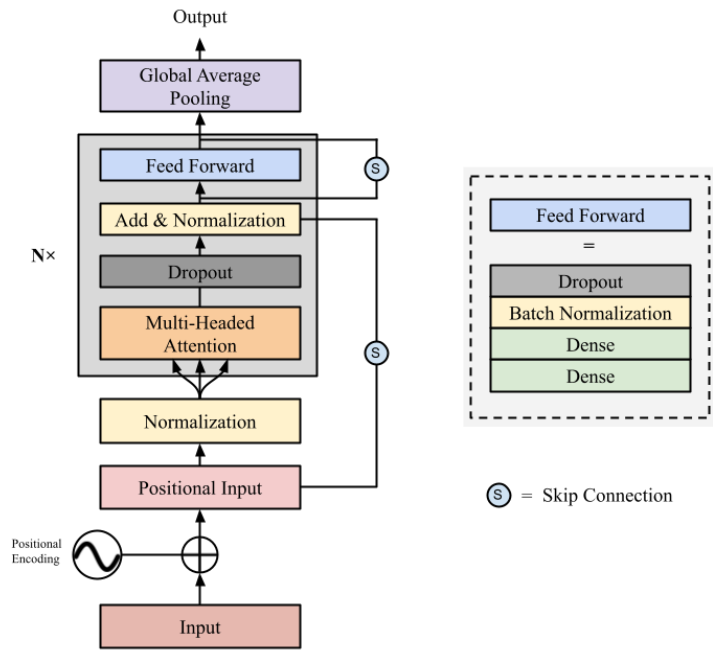

Figure 1: Architecture of the Price-Prediction Encoder-Only Transformer

The evaluation of this price-prediction model involves using Mean Absolute Error (MAE) as the primary evaluation metric, consistent with established practices in time-series forecasting. The choice of MAE is grounded in its relevance and alignment with the objectives of prediction accuracy within this domain. MAE provides a straightforward, interpretable measure of the average magnitude of errors, devoid of directionality, thereby offering a clear quantification of prediction precision. This metric is particularly suited for this context as it encapsulates the absolute deviations between predicted and actual values, offering a direct insight into the model's performance and its practical implications in the field of price prediction. The optimal performance of the model – as evidenced by the lowest Mean Absolute Error (MAE) – was recorded at 0.054 for forecasts targeting price fluctuations within a 30-day horizon. This best-case scenario underscores the model's precision in short-term predictions. To illustrate the dynamic between the temporal distance of the forecast and the model's accuracy (Figure 2), we plotted the average MAE, calculated from three distinct data samples, against the number of months preceding the target prediction date.

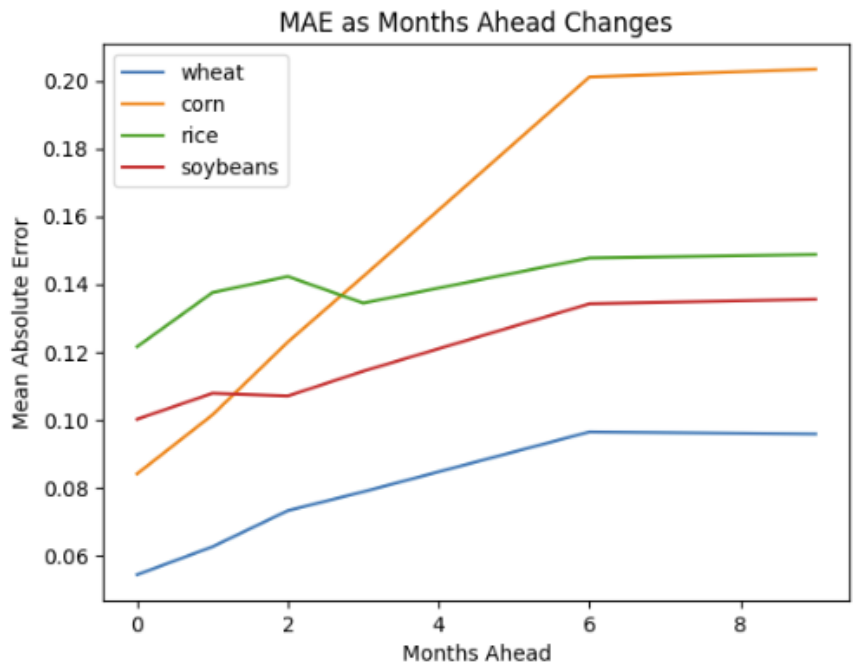

Figure 2: Plot of Average MAE Across Three Samples for Leading Months to Prediction Date

### 3.1.2 Predicting FAO Warning Status

While price prediction serves the purpose of analyzing potential data outcomes, the true value lies in the capability to take actionable steps based on this data, highlighting the importance of a warning system. After our index prediction model started getting a minimal MAE, we transitioned to creating a warning system using the GIEWS data [17]. The project predicts a specific food price while attaching a warning label to a certain time, when appropriate. This warning label can either show no warning label, a moderate warning label, or a high warning label. The placement and severity of the warning labels depend on the data gathered from the GIEWS system [27], specifically growth rates within known countries with similar known commodities. Using the predicted commodity indices, we were able to train our warning model on different commodities and countries. We then used the trained model to output predicted warnings for one specific country and commodity.

The architecture of the transformer used to predict these warnings has a similar architecture to the one used for predicting food commodity prices, outlined in section 3.1.1, 'Predicting Food Commodity Price'. The architecture of the encoder-transformer consists of positional inputs resulting from the input being put through a positional encoding function; a multi-head attention layer along with a dropout and a normalization layer; a section dedicated to adding the layers to the positional inputs; the feed forward network containing multiple dense layers, a batch normalization layer, and an additional dropout layer; a dense layer dedicated to getting the shape back to normal after the feed forward network changes; and a dense layer as the output.

In addition to the standard encoder-transformer, some components are used to increase the model's performance. Specifically, adding L2 regularization in the feed forward network to reduce the over-complexity of the model [28]. Another component is the change of the hidden dimension (layers) in the feed forward network. Throughout the dense layers in the feed forward network, the size of the hidden layers within each dense layer is decreased: this method increases efficiency and helps prevent overfitting. Finally, a skip connection was included to reduce the risk of vanishing gradients and help train the deeper network architecture elements [29].

As the warning model is producing consistent results, the overall score obtained and analyzed is called the F1 score: the F1 score is typically used for multi-classification. F1 score was determined as the most optimal accuracy metric as it accounts for class imbalance issues as there are most labels with no reordered severity. The highest score the model achieved was 72.18% with all the hyperparameter tuning and model architecture modifications. As such, achieving an F1 score of 72.18% indicates that our model, when applied to the given data, aligns with the professional assessments made by the FAO in most cases. This underlines the practical utility of our solution, providing outputs that can be effectively utilized in decision-making processes.

## 3.2 Data Pipeline

Outlined in section 2.1, many data sources are used for the final prediction capabilities presented in this work. To streamline the assimilation of this multifaceted data, we have implemented a comprehensive pipeline encompassing standardized procedures for data cleaning, formatting, and analysis. The methodologies for data cleaning and analysis, as well as formatting, are detailed in sections 2.2 and 4.1, respectively. This systematic approach not only facilitates the extraction of valuable insights from heterogeneous data sources but also empowers future researchers and organizations such as the FAO to efficiently integrate additional variables or features into the analysis. For specific details of this process, including code and methodologies, please refer to the 'data_analysis.ipynb' notebook available on our GitHub repository [22].

## 3.3 Chat-Bot Interface

Commodity food price warnings often lack the necessary context to provide effective guidance, especially for those unfamiliar with economic terminology or market dynamics. To address this challenge, our research harnesses the capabilities of the Llama.cpp [30] large language model (LLM), known for its proficiency in chatbot-related tasks [31] and efficient implementation in C/C++. Consequently, this LLM aims to clarify the significance of these FAO and NourishNet warnings for non-technical users, offering actionable advice in a concise format. The LLM's ability to understand natural language and provide contextually relevant responses makes it an ideal solution for catering to diverse needs and levels of understanding.

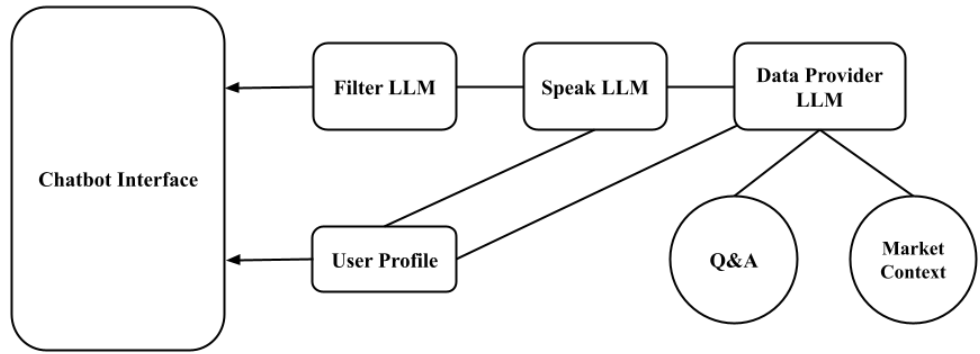

Figure 3: Internal-Interactions Structure of NourishNet Chat-Bot

Within our system architecture, user responses undergo rigorous analysis through a chain-of-thought paradigm, enabling us to determine contextual relevance and outline complex reasoning [32]. Through this chain-of-thought paradigm, there are multiple steps involved with each LLM response, ensuring both accurate and formal conversation. The first step involves response filtration, determining if the user query is irrelevant and subsequently refraining from providing an answer if so (Filter LLM). Conversely, when user inputs pertain to model methodologies, data, or the FAO, the system retrieves relevant responses from a predefined set of question-answer pairs or an embedding space of potential user queries with relevant articles and videos [33]; an effective solution for LLM-knowledge citation to reduce hallucinations. In cases where user inputs do not directly address these specific topics, our LLM generates tailored responses to guide users and provide insights relevant to their queries. For further information about model development, see section 4.3, 'LLM Responsiveness'. Upon examining Figure 3, it is pivotal to acknowledge that the illustration represents a singular Large Language Model (LLM) in operation. Each segment delineates a step in the model's approach to formulating responses, collectively illustrating how the LLM navigates through its chain-of-thought processing.

# 4 Experiments

## 4.1 Formatting Input Data

To optimize data preparation for the encoder-only transformers (sections 3.1.1, 3.1.2), a series of experiments were undertaken. These experiments focused on structuring time-series data in a manner that maximizes information utility while balancing the increased complexity introduced by additional features. This involved the strategic formatting of historical data, encapsulated as previous $mn$ features – where $n$ represents the number of attributes per data point and $m$ denotes the quantity of past samples provided to the model for generating precise forecasts.

The initial strategy entailed organizing the data into conventional 3-dimensional sequence-based structures, specifically of the shape [$num\_samples$, $m$, $n$]. However, these attempts did not yield optimal outcomes when compared to our most effective strategy. The superior approach involved constructing a flattened vector, incorporating only the $m$ most recent prices and futures data. This methodology significantly reduced input dimensionality, quantifiable by a factor of $\frac{m*n}{n-2+2m}$, and enhanced overall model performance.

## 4.2 Transformer Prediction Capabilities

In our comprehensive exploration of the transformer model, a significant portion of the experiments concentrated on hyperparameter optimization. While standard hyperparameters like batch size, optimizer, and epochs were integral to our analysis, we observed that specific hyperparameters had a pronounced impact on model performance. Notably, the transformers detailed in sections 3.1.1 and 3.1.2 were configured with identical hyperparameters, with the exceptions being the number of epochs and transformers. These adjustments were necessary to balance data complexity, prevent overfitting, and ensure the model's capacity to comprehend the dataset effectively.

**Key attributes of our optimized hyperparameter configuration include:**

- **L2 regularization** was set to 0.003 to mitigate the risk of overfitting by penalizing large weights, thereby encouraging the model to learn more generalized patterns that are not overly dependent on the training data.
- $m$ was made 3 to match the temporal structure of fiscal quarters, ensuring that the model captures the cyclicality inherent in quarterly financial data. This alignment with fiscal quarters is particularly relevant for the futures data, which is contractually bound to quarter periods.
- **Dropout rate** of 0.4 was selected after empirical testing suggested that a higher rate was beneficial in preventing the model from becoming overly reliant on specific nodes, thus enhancing its generalization capabilities. This rate is particularly effective in introducing a level of randomness during training, which helps the model to avoid overfitting.
- **Batch size** was chosen as 3 to reflect the quarterly temporal resolution of the data. This size helps the model to effectively learn and adapt to the nuances of quarter-based trends, ensuring that each batch offers a representative snapshot of the underlying temporal dynamics [34].

These hyperparameter settings were derived from a rigorous experimentation process, aiming to fine-tune the model for optimal performance on the task of understanding and predicting based on the provided dataset. This approach mirrors the process undertaken to finalize the model's architecture, detailed in sections 3.1.1 and 3.1.2. Through iterative testing and validation, we fine-tuned these parameters to achieve a delicate balance, ensuring the model's adeptness at capturing the dataset's nuances without succumbing to overfitting or underfitting, thereby solidifying the foundation for our model's predictive accuracy and reliability.

### 4.3 LLM Responsiveness

In our research, we introduce an innovative communication framework tailored for LLMs, focusing on guiding users through commodity food price warnings. Unlike the widely used LangChain [35] method, our framework emphasizes a chain-of-thought [36] structure designed specifically for LLM interactions. This approach fosters a dynamic conversation flow, where user inputs are analyzed within the ongoing dialogue context. By employing a chain-of-thought paradigm, our framework ensures the LLM can effectively track conversation trajectory and respond relevantly and coherently. This iterative process encourages deeper user engagement and understanding, especially in navigating complex topics like commodity food price warnings. Moreover, the adaptability of this approach allows the LLM to adjust responses based on evolving user queries and contextual nuances, ultimately enhancing user experience, and providing actionable advice tailored to individual needs. Thus, the chain-of-thought process forms a crucial element of our framework, enabling optimal performance and adaptability in diverse user interactions.

When designing the LLM, multiple configurations were taken into consideration. The initial configuration serves as an input to generate the initial conversation produced by the LLM before user interaction. This system excels in responding optimally to the current context, which involves introducing the country, severity, commodity, and language, as it functions as a chatbot for the FAO. The filter configuration engages with the LLM to assess the relevance of the prompt and determine whether it warrants a response. It evaluates the appropriateness of prompts and decides whether they should be filtered, responding accordingly. The user profile configuration establishes a distinct profile utilized between each message, enhancing memory retention, and facilitating the recollection of crucial information from previous prompts. This profile encompasses essential details such as commodity, country, severity, and other pertinent information.

# 5 Challenges

## 5.1 FAO Warning Labels

To compile our warning dataset, we engaged in a detailed review of the FAO's monthly reports [17] from 2015 to 2024, due to the unavailability of a consolidated public database on food scarcity warnings. This meticulous process was necessary to ensure the integrity and comprehensiveness of our dataset. While this endeavor was resource-intensive, it was instrumental in creating a robust foundation for our research. We acknowledge the efforts of our research team in minimizing potential errors during data collection to maintain the dataset's accuracy and reliability.

In 2015, the FAO enhanced the format of its food warning reports, offering more detailed statistics, which significantly benefitted our analysis. Consequently, we focused on utilizing the reports post-2015 for our dataset, ensuring we leveraged the most detailed and relevant information available for our study period up to 2024. This approach not only aligns with our commitment to precision but also reflects our adaptive strategy in response to evolving data quality and availability.

## 5.2 Data Overlap Scarcity

Addressing data overlap scarcity is crucial for ensuring the reliability and accuracy of our model. When dealing with multiple data sources, the potential for overlap is low, but ensuring that each data sample is present across all sources helps maintain consistency and completeness in the training dataset [37]. This consistency is vital because it reduces the risk of bias in the model's predictions. If certain data samples were missing from one or more sources, it could lead to gaps in the model's understanding of certain country commodity pairs, potentially resulting in inaccurate predictions or biased outcomes. By carefully curating the training data to include only complete and consistent samples, we can improve the robustness and generalizability of our AI models, making them more effective tools for forecasting food commodity prices and warning statuses. Avoiding sparse datasets helps prevent the introduction of noise or uncertainty into the model, further enhancing the data's reliability, consequently increasing the model's performance.

# 6 Future Work

## 6.1: Modeling Nations as RL Agents

Achieving consistency in predicting specific food indices globally and issuing warning labels over time can significantly benefit individuals and businesses, enabling them to make more informed decisions. However, merely determining price and change levels represents only one aspect of the broader decision-making process. The other crucial component involves utilizing this information to make financial decisions, whether from an individual or national perspective.

In a relevant paper [38] focusing on the Foreign Exchange Market (Forex) and commodity markets, which are among the largest markets globally, the authors proposed a more efficient approach to automated profit generation. By leveraging Deep Reinforcement Learning, they trained and tested an agent in a time-series environment to execute trades on market data within a simulated setting. Inspired by such advancements, our future endeavors aim to develop an alert system capable of identifying optimal times for buying, selling, or holding specific commodities for various nations. This system would empower stakeholders with timely insights, enabling them to capitalize on market opportunities and mitigate risks effectively.

### 6.2: Developing Generative AI Training Environment

Developing a specialized training environment for Generative AI tailored to the complexities of commodity food price warning predictions offers a promising avenue for enhancing our approach. Leveraging synthetic data generated through sophisticated techniques like Multivariate Time Series Simulation Generative Adversarial Networks (MTSS-GAN) [39] is particularly well-suited for our solution, given its capability to understand multiple variables and their intricate connections within time series data. This augmentation allows our models to learn from a broader range of scenarios, including rare events and extreme market conditions, crucial for robust warning prediction systems. Given that time series data for commodities is recorded only monthly, the inherent limitation in the quantity of historical data highlights the importance of synthetic data generation. Synthetic data provides a means to expand the dataset, enabling more comprehensive training and enhancing the predictive capabilities of our models. Increasing the volume of data available for model training not only facilitates better understanding of underlying patterns but also enhances the model's ability to generalize to unseen data, thereby improving the accuracy and reliability of commodity food price warning predictions [40].

## 7 Conclusion

This research introduces NourishNet, a sophisticated pipeline designed for aggregating and analyzing data relevant to food commodities from an array of sources and formats, achieving notable advancements in the domain of monthly price prediction. Central to NourishNet is its alert system, which demonstrates a 72.18% alignment rate (F1-Score) with the professional warnings issued by the Food and Agriculture Organization (FAO). The inclusion of a multilingual chat interface serves to lower barriers to entry, enabling a broader range of users to access, understand, and respond to critical information pertinent to various commodities and geographical regions. NourishNet's development is primarily aimed at enhancing decision-making processes within the global food supply chain, particularly to aid underdeveloped communities that depend on FAO's alerts to preempt food price hikes and potential shortages. Therefore, the continual collaboration with the FAO represents a significant stride in proposing a deep learning-based tool to augment traditional statistical approaches, marking a pivotal contribution to the application of artificial intelligence in addressing critical issues within the food supply chain.